\pdfoutput=1
\documentclass{article}

\usepackage{microtype}
\usepackage{graphicx}
\usepackage{subfigure}
\usepackage{booktabs} % for professional tables

\usepackage{hyperref}
\usepackage[accepted]{icml2020}

\usepackage{amsmath,amsfonts,bm}

\def\eqref#1{equation~\ref{#1}}
\def\1{\bm{1}}

\def\rvx{{\mathbf{x}}}
\def\rvy{{\mathbf{y}}}

\def\vx{{\bm{x}}}
\def\vy{{\bm{y}}}

\DeclareMathAlphabet{\mathsfit}{\encodingdefault}{\sfdefault}{m}{sl}
\SetMathAlphabet{\mathsfit}{bold}{\encodingdefault}{\sfdefault}{bx}{n}

\def\sY{{\mathbb{Y}}}

\usepackage{url}
\usepackage{bbm}
\usepackage[inline]{enumitem}
\usepackage{multirow}
\usepackage{float}
\usepackage{nicefrac}

\def\py{p_{\rvy}}

\def\pyx{p_{\rvy \mid \rvx}}
\def\pxy{p_{\rvx \mid \rvy}}

\def\qyx{q_{\rvy \mid \rvx}}
\def\ryx{r_{\rvy \mid \rvx}}
\def\bpyx{\bar{p}_{\rvy \mid \rvx}}

\newcommand{\hl}[1]{\textbf{#1}}
\newcommand{\red}[1]{{\color{red}#1}}

\begin{document}

\twocolumn[
\icmltitle{Semi-Supervised Speech Recognition via Local Prior Matching}



\icmlsetsymbol{equal}{*}

\begin{icmlauthorlist}
\icmlauthor{Wei-Ning Hsu}{mit} % how to add "work performed during..."?
\icmlauthor{Ann Lee}{fair}
\icmlauthor{Gabriel Synnaeve}{fair}
\icmlauthor{Awni Hannun}{fair}
\end{icmlauthorlist}

\icmlaffiliation{mit}{Massachusetts Institute of Technology}
\icmlaffiliation{fair}{Facebook AI Research}

\icmlcorrespondingauthor{Wei-Ning Hsu}{wnhsu@mit.edu}
\icmlcorrespondingauthor{Awni Hannun}{awni@fb.com}

\icmlkeywords{Speech Recognition, Semi-Supervised Learning, Knowledge Distillation}

\vskip 0.3in
]



\printAffiliationsAndNotice{}  % leave blank if no need to mention equal contribution

\begin{abstract}
For sequence transduction tasks like speech recognition, a strong structured prior model encodes rich information about the target space, implicitly ruling out invalid sequences by assigning them low probability. In this work, we propose local prior matching (LPM), a semi-supervised objective that distills knowledge from a strong prior (e.g. a language model) to provide learning signal to a discriminative model trained on unlabeled speech. We demonstrate that LPM is theoretically well-motivated, simple to implement, and superior to existing knowledge distillation techniques under comparable settings. Starting from a baseline trained on 100 hours of labeled speech, with an additional 360 hours of unlabeled data, LPM recovers 54\% and 73\% of the word error rate on clean and noisy test sets relative to a fully supervised model on the same data.\footnote{Code and models are available at \url{https://github.com/facebookresearch/wav2letter/tree/master/recipes/models/local_prior_match}}
\end{abstract}

\section{Introduction}
Fully supervised learning remains the mainstream paradigm for state-of-the-art automatic speech recognition (ASR). These systems require huge annotated data sets~\citep{li2017acoustic, chiu2018state, hannun2014deep, amodei2016deep},
which are time-consuming and expensive to collect. This hinders the development of accurate ASR for low resource languages~\citep{precoda2013non}.
In fact, out of over 6,000 spoken languages, fewer than 150 are supported by commercial ASR service providers.
In sharp contrast to how we teach machines to recognize speech, humans do not learn by listening to thousands of hours of speech and simultaneously reading the corresponding transcriptions. 
Instead, humans possess an inherent ability to learn from vast quantities of unlabeled speech~\citep{chomsky1986knowledge,kuhl2004,glass2012towards,dupoux2018cognitive}.
Consider conversing with someone with a strong accent. Even when the speaker pronounces several words in an unusual way, one can often correctly understand the sentence.
We argue that the source of indirect supervision in processing unlabeled speech comes from prior knowledge about the world and the context of the speech.

Inspired by this, we devise a semi-supervised learning framework termed \emph{local prior matching} (LPM). We apply LPM to speech recognition allowing an ASR model to learn from unlabeled speech by leveraging a strong language model.
Given an unlabeled utterance, a proposal model first generates a set of hypotheses. The language model (LM) then produces a target distribution for the ASR model to match, enabling distillation of prior knowledge into the ASR model.

We evaluate LPM on the LibriSpeech corpus~\citep{panayotov2015librispeech}, using 100 hours of labeled speech to build a baseline.
With the addition of 360 hours of unlabeled data, LPM recovers 54\% and 73\% of the word error rate (WER) on a clean and noisy test set relative to a completely supervised model on the full 460 hours.
By augmenting LPM with another 500 hours, for a total of 860 hours of unlabeled speech, LPM surpasses the performance of using 460 hours of labeled data.
We also conduct extensive ablation studies in order to demonstrate the significance of each proposed component.
Our main contributions are as follows:

\begin{itemize}
    \item We propose an intuitive yet theoretically well-motivated learning objective that can leverage large quantities of unpaired speech and text.
    \item We achieve a state-of-the-art result in WER recovery with unlabelled data on a standard ASR benchmark.
    \item We show that LPM can scale to 60,000 hours of unlabelled speech and yield further gains in WER.
\end{itemize}    

Compared to adversarial training~\cite{liu2019adversarial}, back-translation~\cite{hayashi2018back}, and cycle-consistency~\cite{hori2019cycle}, LPM achieves significantly better performance without the need to jointly train additional modules. Compared to knowledge distillation, (1) LPM distills from a prior rather than a posterior, (2) considers multiple hypotheses in a principled manner and (3) improves the proposal model which results in improved WER. We show that LPM achieves better WER compared to a strong pseudo-label baseline~\cite{kahn2019self} as well as other forms of knowledge distillation.
\section{Method}

\subsection{Preliminaries}
Let $\rvx$ denote an utterance of speech and $\rvy$ a transcription. We assume speech is generated following a two-step process:
\begin{equation}
    \rvy \sim \py
    \quad\quad\quad\quad
    \rvx \sim \pxy,
\end{equation}
where the text $\rvy$ is first generated from the language model (LM), $\py$, and the speech $\rvx$ is then generated from a text-to-speech (TTS) model, $\pxy$, conditioned on $\rvy$.
The posterior, $\pyx$, is then the ASR model of interest.
In a typical supervised learning setting, one has access to a labeled dataset $\mathcal{D}_{l}$, which contains paired samples $\{ (\vx^{(i)}, \vy^{(i)}) \}_{i=1}^N$ drawn from the joint distribution, $p_{\rvx\rvy}(\rvx, \rvy)$.
An ASR model $\qyx$ can be trained by minimizing the marginal weighted cross entropy $\mathbb{E}_{p_\rvx}[ H( p_{\rvy \mid \rvx}, q_{\rvy \mid \rvx} ) ]$, which can be estimated with $\sum_{(\vx, \vy) \in \mathcal{D}_{l}} -\log \qyx(\vy \mid \vx)$ from samples. 

In the semi-supervised learning setting, we have additional unpaired speech $\mathcal{D}_u^{s}$ and text $\mathcal{D}_u^{t}$, both of which can be many times larger than the paired dataset $\mathcal{D}_l$. We wish to exploit this unpaired data to improve the ASR model. To that end, we propose a method to estimate $p_{\rvy \mid \rvx}$ for an unlabeled example $\vx$.

\subsection{Through the Lens of Generative Modeling}

A natural way to approximate the posterior, $\pyx$, is to estimate a TTS model, $\pxy$, from paired data, an LM, $\py$ from unpaired text and apply Bayes' theorem
\begin{equation}
\label{eq:bayes}
    \pyx(\vy \mid \vx) 
    = \dfrac{p_\rvy(\vy)\;\pxy(\vx \mid \vy)}
    {\sum_{\hat{\vy} \in \sY} p_\rvy(\hat{\vy})\;\pxy(\vx \mid \hat{\vy})}.
\end{equation}
However, for sequence transduction tasks, the cardinality of the output space $\sY$ is infinite hence marginalization is intractable.
Luckily, the denominator in equation~\ref{eq:bayes} can often be approximated by summing over a set of hypotheses proposed by a trained ASR model, as done in sequence discriminative training~\cite{povey2005discriminative,vesely2013sequence} and differentiable beam search decoding~\cite{collobert2019fully}.
Such an approximation is reasonable because only text sequences that are linguistically and acoustically plausible will contribute non-negligible probability in the marginalization, and there are very few of them for a given utterance.

\begin{table*}[t]
    \caption{Beam search hypotheses from the supervised model (\emph{Sup.}) trained on 100hr $\mathcal{D}_{l}$ and the LPM model (\emph{LPM}) trained on 100hr $\mathcal{D}_{l}$ and 360hr $\mathcal{D}_u^s$. Hypotheses are ranked by the ASR score $\qyx$. The reference transcription (\emph{Ref.}) is also shown for each example.}
    \label{tab:decsample}
    \begin{center}
    \resizebox{\textwidth}{!}{
    \begin{tabular}{lccl}
        \toprule
        
        Model & Rank & $\log \py$ & Beam Search Hypotheses ($k=4)$ \\
        \midrule\midrule
        Ref. & - & -34.14 & \_she \_walk ed \_very \_fast \_after \_she \_left \_the \_house \$ \\
        \midrule
        \multirow{4}{*}{Sup.}
        & 1 & -47.58 & \_she \red{\_looked} \_very \red{\_thought} \_after \_she \_left \_the \_house \$ \\
        & 2 & -38.54 & \_she \red{\_looked} \_very \red{\_fat} \_after \_she \_left \_the \_house \$ \\
        & 3 & -53.47 & \_she \red{\_what} \_very \red{\_thought} \_after \_she \_left \_the \_house \$ \\
        & 4 & -83.85 & \_she \red{\_ w o u l t} \_very \red{\_thought} \_after \_she \_left \_the \_house \$ \\
        \midrule
        \multirow{4}{*}{LPM}
        & 1 & -32.24 & \_she \_walk ed \_very \_fast \red{\_as} \_she \_left \_the \_house \$ \\
        & 2 & -34.14 & \_she \_walk ed \_very \_fast \_after \_she \_left \_the \_house \$ \\
        & 3 & -33.10 & \_she \red{\_looked} \_very \_fast \red{\_as} \_she \_left \_the \_house \$ \\
        & 4 & -36.59 & \_she \red{\_looked} \_very \_fast \_after \_she \_left \_the \_house \$ \\
        \midrule\midrule
        Ref. & - & -68.01 & 
        \_oh \_if \_i \_had \_imagined \_him \_still \_in \_such \_distress
        \_sure ly \_i \_might \_have \_done \_something \_to \_help \_him \$ \\
        \midrule
        \multirow{4}{*}{Sup.}
        & 1 & -110.11 & 
        \red{\_i \_before} \_i \_had \_imagined \_him \red{\_steal ing} \_such \_distress 
        \_sure ly \_i \red{\_why} \_have \_done \_something \_to \_help \red{\_you} \$ \\
        & 2 & -107.55 & 
        \red{\_i \_before} \_i \_had \_imagined \_him \red{\_steal ing} \_such \_distress 
        \_sure ly \_i \red{\_want} \_have \_done \_something \_to \_help \red{\_you} \$ \\
        & 3  & -107.81 & 
        \red{\_i \_before} \_i \_had \_imagined \_him \_still \red{ing} \_such \_distress 
        \_sure ly \_i \red{\_want} \_have \_done \_something \_to \_help \red{\_you} \$ \\
        & 4  & -107.10 & 
        \red{\_i \_before} \_i \_had \_imagined \_him \red{\_steal ing} \_such \_distress 
        \_sure ly \_i \red{\_want \_of} \_done \_something \_to \_help \red{\_you} \$ \\
        \midrule
        \multirow{4}{*}{LPM}
        & 1 & -72.55 & 
        \_oh \_if \_i \_had \_imagined \_him \_still \_in \_such \_distress 
        \_sure ly \_i \red{\_would} \_have \_done \_something \_to \_help \red{\_you} \$ \\
        & 2 & -71.35 & 
        \_oh \_if \_i \_had \_imagined \_him \_still \_in \_such \_distress 
        \_sure ly \_i \_might \_have \_done \_something \_to \_help \red{\_you} \$ \\
        & 3 & -69.29 &  
        \_oh \_if \_i \_had \_imagined \_him \_still \_in \_such \_distress 
        \_sure ly \_i \red{\_would} \_have \_done \_something \_to \_help \_him \$ \\
        & 4 & -85.62 & 
        \_oh \_if \_i \_had \_imagined \_him \_still \_in \_such \_distress 
        \_sure ly \_i \red{\_won't} \_have \_done \_something \_to \_help \red{\_you} \$ \\
        \bottomrule
        
    \end{tabular}
    }
    \end{center}
\end{table*}

\subsection{Local Prior Matching}
Let $B(\vx)$ be the set of beam search hypotheses generated by a proposal ASR model $\ryx$ with a beam size of $k$. By replacing $\sY$ with $B(\vx)$, our estimated posterior becomes:
\begin{equation}
    \bpyx(\vy \mid \vx) 
    = \dfrac{p_\rvy(\vy)\;\pxy(\vx \mid \vy)}
    {\sum_{\hat{\vy} \in B(\vx)} p_\rvy(\hat{\vy})\;\pxy(\vx \mid \hat{\vy})}  \; \mathbbm{1}(\vy \in B(\vx)) , \nonumber
\end{equation}
where $\mathbbm{1}(\cdot)$ is the indicator function used to ensure $\bpyx$ is a valid distribution.

Table~\ref{tab:decsample} shows the reference text (\emph{Ref.}) and the hypotheses generated by a supervised ASR model trained on 100 hours of paired data (\emph{Sup.}) for two utterances. We make three key observations. 
\begin{enumerate*}[label=(\arabic*)]
    \item The hypotheses are acoustically similar to each other (e.g., \emph{``stealing''} / \emph{``still ing''}), indicating that the acoustic probability $\pxy$ between the hypotheses may be very close. 
    \item One can often tell which hypotheses are wrong without listening to the speech because they are semantically unreasonable and grammatically incorrect in multiple locations.
    \item The third column shows the linguistic scores $\log \py$ from an LM trained on unpaired text $\mathcal{D}_u^{t}$. Within each utterance, linguistic scores align well with linguistic plausibility.
\end{enumerate*} 

Based on the above observations, we assume that the posterior probability between hypotheses are dominated by $\py$, whereas $\pxy$ can be treated as a constant.
Therefore, our final posterior approximation can be written as 
\begin{equation}
    \bpyx(\vy \mid \vx) 
    = \dfrac{p_\rvy(\vy) }
    {\sum_{\hat{\vy} \in B(\vx)} p_\rvy(\hat{\vy})} \; \mathbbm{1}(\vy \in B(\vx)) . \label{eq:locprior}
\end{equation}
The approximated posterior only requires computing language model probabilities of the beam search hypotheses.
We refer to \eqref{eq:locprior} as the \emph{local prior}, since it is the prior re-normalized with intended support only in the neighborhood of the unknown target transcription.
We also propose \emph{local prior matching} (LPM) as a semi-supervised objective for training an ASR model $\qyx$ with unlabeled speech $\vx$:
\begin{align}
    &\mathcal{L}_{lpm} (\qyx; \vx, \py, \ryx, k) 
    = H(\bar{p}_{\rvy \mid \rvx=\vx}, q_{\rvy \mid \rvx=\vx}) \nonumber\\
    &=- \sum_{\vy \in B(\vx)} \dfrac
        {  p_\rvy(\vy) }
        { \sum_{\hat{\vy} \in B(\vx)} 
        p_\rvy(\hat{\vy}) }
    \log\qyx(\vy \mid \vx). \nonumber
\end{align}
The LPM objective minimizes the cross entropy between the local prior and the model distribution, and is minimized when $\qyx = \bpyx$.
Intuitively, LPM encourages the ASR model to assign posterior probabilities proportional to the linguistic probabilities of the proposed hypotheses, similar to how humans recognize speech when ambiguity exists (e.g., \emph{``let her''} / \emph{``led her''} / \emph{``letter''}).
For clarity, we term $\qyx$ the \emph{online model}, and let $\qyx(\rvy \mid \rvx;\; \theta_q)$ and $\ryx(\rvy \mid \rvx;\; \theta_r)$ denote the models with parameters $\theta_q$ and $\theta_r$, respectively.

\subsection{Choice of Proposal Model}
\label{sec:offpolicy}

The quality of the posterior approximation $\bpyx$ depends on the proposal model $\ryx$.
Instead of using a fixed proposal model throughout the entire training process, we consider two strategies for updating $\ryx$ with $\qyx$.

\textbf{On-policy beam search}\hspace{.3cm} The first approach always uses the online model $\qyx$ as the proposal model. This means $\theta_r=\theta_q$ and is effectively a form of \emph{on-policy beam search}, since the model used to generate hypotheses is also the model we update.

\textbf{Off-policy beam search}\hspace{.3cm} While the on-policy method benefits from the immediate improvement of the online model, it also suffers immediately if the gradient update from a mini-batch deteriorates performance. This can result in instability during optimization.
We consider a second option which does not tie $\theta_r$ and $\theta_q$ but instead updates $\theta_r$ with $\theta_q$ every $T$ steps only when the performance of the online model $\qyx$ is better than that of the proposal model $\ryx$ by some metric.
We refer to the second option as \emph{off-policy beam search}.
To avoid overfitting to the training set, we use the character error rate (CER) on the validation set as the metric for the proposal model update.
We set $T = 1000$ for all experiments with the off-policy beam search.

\subsection{Filtering Hypotheses Using Estimated Lengths}\label{sec:filter}
As noted in \citet{chorowski2016towards}, sequence-to-sequence ASR models sometimes predict end-of-sentence (EOS) tokens too early or generate looping n-grams, resulting in hypotheses that are significantly shorter or longer than the set of acoustically matched texts for a given utterance.
Of the two failure modes, the former is more harmful when using LPM. The reason is that the LPM objective assumes all hypotheses obtained from the beam search are acoustically reasonable, and weights each of them by linguistic plausibility given by an LM.
While LMs are effective in discriminating plausibility between sentences of similar length, we find empirically they tend to assign higher probabilities to shorter sentences than to longer sentences, even when the longer ones are more plausible and grammatically correct than the shorter ones.
As a result, truncated hypotheses are assigned higher weights than acoustically matched but longer ones, which in turn encourages earlier prediction of EOS tokens and forms a catastrophic feedback loop particularly with the on-policy beam search.

To address this issue, we propose a simple filtering heuristic based on the text length.
Before training the model, a text length $L$ is estimated for each unlabeled speech sample $\vx$. 
During training, only hypotheses with length close to $L$ are retained for the LPM objective computation.
Let $len(\vy)$ denote the length of $\vy$.  We keep a hypothesis $\vy$ only if $\lfloor r_{lb} \cdot L \rfloor \le len(\vy) \le \lceil r_{ub} \cdot L \rceil$, where $r_{lb}$ and $r_{ub}$ are the text length lower and upper bound ratios, respectively.
Several methods can be used to estimate the text length on an unlabeled utterance, including the average speaking rate~\citep{peng2019parallel} or a phoneme/syllable segmentation~\citep{adell2004towards, scharenborg2010unsupervised, wang2017gate}.
In this work, we estimate the length by using using that of the best hypothesis generated from the initial proposal model, generated with either ASR-only greedy decoding or ASR+LM beam search decoding. 
\section{Related Work}\label{sec:related}
Our work builds on a large body of work in semi-supervised learning for ASR.
Research in this direction can be classified based on the required modules and the objectives used to learn from unpaired data.

In \citet{drexler2018combining} and \citet{karita2018semi}, bi-encoder network architectures are used, which map text and speech to representations in a shared space with their corresponding encoder, and then apply a shared decoder to map from the shared space to the text space.
Another line of work adds a TTS model~\citep{tjandra2017listening, tjandra2019end, baskar2019self} or a text-to-encoding (TTE) model~\citep{hayashi2018back, hori2019cycle} in the loop of ASR training, which can be utilized for back-translation style data augmentation~\citep{sennrich2015improving} or cycle-consistency training~\citep{zhu2017unpaired}.
\citet{liu2019adversarial} treats ASR as a generative model that conditions on speech instead of random noise vectors, and adopts the generative adversarial network (GAN)~\citep{goodfellow2014generative} framework in order to improve the fidelity of ASR-generated texts.
All the aforementioned methods involve additional modules that must be jointly optimized with the ASR model and require finding a careful balance between multiple training objectives. In contrast, LPM only requires a pre-trained LM and optimizes a principled cross-entropy objective.

Knowledge distillation (KD)~\cite{cui2017knowledge, parthasarathi2019lessons} and weak distillation (also known as \emph{self-training} or \emph{pseudo-labeling})~\cite{vesely2017semi, manohar2018semi, li2019semi, kahn2019self} have also achieved great success in semi-supervised learning for ASR.
In KD, a student posterior model learns from a teacher posterior model by minimizing the cross entropy between the two distributions on the unlabeled speech data. Because of this, we expect KD to yield better student models when when the teacher distribution starts out better than that of the student.
On the other hand, in weak distillation the teacher distribution is replaced with its mode. Hence, weak KD is equivalent to training on the unlabeled data with a supervised maximum likelihood objective, using the labels predicted by the teacher model. 
In this case, the teacher model can be the same as the student model, a case commonly known as self-training or pseudo-labelling. To obtain pseudo-labels with higher quality, LMs are used for shallow fusion decoding~\cite{chorowski2016towards}, which requires expensive hyperparameter search on a held-out set.
This can be viewed as interpolating between an estimated prior and posterior to obtain a better teacher model to distill from.

We can view LPM as a type of knowledge distillation, but with three key differences.
First, LPM distills directly from a prior instead of a posterior, enabling seamless integration of available context (e.g., the previous sentence or other modalities).
Second, LPM considers multiple hypotheses and provides a principled way to weight them, while weak distillation typically uses only one hypothesis~\cite{li2019semi} or assumes a uniform distribution when multiple hypotheses are used~\cite{kahn2019self}.
Third, LPM uses an improving proposal model with a stable update strategy. This is difficult to implement with pseudo-labels generated with an LM because the hyperparameters used in decoding should be updated as the teacher changes.
We demonstrate the significance of these differences in our experiments.

Aside from semi-supervised learning, this work is also related to unsupervised domain adaptation, where unlabeled speech of the target domain is provided.
Unlike the proposed method, previous studies focus on learning domain invariant features~\citep{sun2017unsupervised, hsu2018extracting, meng2017unsupervised, meng2018adversarial, meng2019attentive} or data augmentation through learned speech transformations~\citep{hsu2017domain, hsu2018unsupervised}.
Our work also shares a similar motivation as posterior regularization (PR)~\cite{ganchev2010posterior}. Both methods aim to incorporates prior knowledge to improve a posterior, though PR achieves this by adding handcrafted linear constraints with a limited family of posterior distributions.
\section{Experimental Setup}
\paragraph{Dataset} We evaluate our approach on LibriSpeech~\citep{panayotov2015librispeech}, a crowd-sourced audio book corpus derived from the \citet{LibriVox}.
The training set contains 960 hours of speech, officially split into three sets: train-clean-100, train-clean-360, and train-other-500, where the first two sets are easier and the third set is noisier and more accented. 
Similarly, the development and test sets are also split according to difficulty, resulting in four partitions: \{dev, test\}$\times$\{clean, other\}, each of which contains roughly five hours of speech.
In this work we use train-clean-100 as the paired speech data, and the other two training splits as the unpaired data. We also examine how well LPM scales to a much larger amount of unlabelled data. To do this, we use the recently released Libri-Light corpus~\cite{kahn2019libri} which contains roughly 60k hours of unlabelled audio from the same domain as LibriSpeech.

We train the LM on the unpaired text data provided with LibriSpeech, which includes approximately 14,500 books collected from \citet{Gutenberg}.
Some of the books in the text corpus overlap with those in the LibriSpeech training set.
To avoid training the LM on the ground truth text of the unlabeled speech, we exclude the 997 overlapping books from the text data.
We follow the same recipe as~\citet{kahn2019self} to pre-process the remaining text.

\paragraph{Neural Network Architecture} 
The proposal model $\ryx$ and the online model $\qyx$ are sequence-to-sequence neural networks~\citep{bahdanau2016end, chorowski2016towards} with the same time-depth separable (TDS) architecture proposed in~\citet{hannun2019sequence}. 
The encoder is fully convolutional, composed of TDS blocks which reduce the number of parameters while keeping the receptive field large.
The decoder is a single layer recurrent neural network (RNN) with gated recurrent units (GRUs), equipped with a single-headed inner-product key-value attention~\citep{vaswani2017attention} for querying information from the encoder outputs.
Unless otherwise stated, we follow the recipe of~\citet{kahn2019self} which uses fewer TDS blocks in the encoder compared to ~\citet{hannun2019sequence} in order to generalize better when trained on the smaller LibriSpeech train-clean-100.
The output of the decoder at each step is a posterior distribution over 5,000 word pieces. The word pieces are generated with the SentencePiece toolkit~\citep{kudo2018sentencepiece} using transcripts from train-clean-100.

To enable efficient evaluation of the language model probabilities, which is required at each training step, we use the gated convolutional language model architecture (ConvLM) proposed in~\citet{dauphin2017language}, which achieves competitive performances compared to recurrent models while significantly reducing the latency.
We use the same 5,000 word-piece vocabulary for the LM which is trained with the same model configuration and recipe as~\citet{zeghidour2018fully}.
The trained ConvLM achieves a token perplexity of 34.24 on the development set.

\paragraph{Optimization}
We use both paired and unpaired data to optimize $\qyx$. To simplify the optimization procedure, the model is provided with either a paired or an unpaired batch at each step, alternated with a fixed ratio $m_l : m_u$.
When given a paired batch of $n$ samples, $\{(\vx^{(i)}, \vy^{(i)})\}_{i=1}^{n}$, the model minimizes the standard cross-entropy loss, $\frac{1}{n}\sum_{i} -\log \qyx(\vy^{(i)}~\mid~\vx^{(i)})$.
When provided with an unpaired batch $\{ \vx^{(i)} \}_{i=1}^{n}$, the model minimizes a weighted LPM loss, $\frac{\alpha}{n} \cdot \sum_{i} \mathcal{L}_{lpm}(\qyx; \vx^{(i)}, \py, \ryx, k)$. The weight $\alpha$ and the mixing ratio $m_l : m_u$ are used to balance the supervised and self-training objectives.
For regularization we use 20\% dropout~\citep{srivastava2014dropout}, 10\% label smoothing, 1\% decoder input sampling, and 1\% word piece sampling~\citep{kudo2018subword} following~\citet{kahn2019self}.
We use SGD without momentum to train the online model with an initial learning rate of 5e-2.
To achieve a good CER on the development sets, the model is trained for at least 1.6M steps (paired and unpaired) with a batch size of 16 (8 GPUs $\times$ 2 per GPU).
The learning rate is annealed by a factor of two every 0.64M steps.
All experiments in this paper are implemented in the wav2letter++ framework~\citep{pratap2018wav2letter++}.

\paragraph{Initialization}
To initialize the proposal model and the online model, we consider three checkpoints from a baseline model trained on train-clean-100 for a varying number of steps using only the supervised objective.
The three checkpoints, whose parameters are denoted as $\theta_A$, $\theta_B$, and $\theta_C$, are trained for about 300k / 40k / 16k steps, achieving average development set CERs of 13\% / 20\% / 38\%, respectively.
We hypothesize that initializing the proposal model from the best checkpoint leads to a better approximation $\bpyx$ to the posterior. 
In contrast, \citet{kahn2019self} observe that training from scratch achieves consistently better performance than starting from a well-trained model, thus we hypothesize that initializing the online model with an earlier checkpoint may lead to better results.
We initialize $\theta_r = \theta_A$ and $\theta_q = \theta_C$ unless otherwise specified.
\section{Results}

The best supervised model trained only on train-clean-100 ($\theta_A$) achieves a (no LM) WER of 14.00\%/37.02\% on dev-clean/dev-other, respectively.
Unless otherwise stated, we use train-clean-360 as the unpaired speech dataset, $(r_{lb}, r_{ub}) = (0.95, 1.05)$ for length filtering with reference lengths obtained from ASR-only greedy decoding.

\subsection{Beam Size, Mixing Ratio, and LPM Weights}

\begin{table*}[ht]
    \small
    \caption{We vary the mixing ratio $m_l : m_u$ and the beam size $k$. An LPM weight $\alpha\!=\!0.2$ is used.}
    \label{tab:beam}
    \begin{center}
    \begin{tabular}{llccccc}
        \toprule
        \multirow{2}{*}{$\mathcal{D}_u^s$} & \multirow{2}{*}{$m_l : m_u$} & 
        \multicolumn{5}{c}{dev-clean / dev-other WER (\%)} \\
        & & $k=1$ & $k=2$ & $k=4$ & $k=8$ & $k=16$ \\ 
        \midrule
        \multirow{4}{*}{360hr}
        & $4:1$ & 10.75 / 31.62 & 10.60 / 30.96 & 10.25 / 30.67 & 10.14 / 30.17 & 10.09 / 29.99 \\
        & $1:1$ & 10.43 / 29.76 &  9.56 / 28.83 &  9.37 / 28.10 &  9.06 / 27.35 &  8.88 / 27.25 \\
        & $1:4$ & 11.09 / 29.89 &  9.34 / 27.45 &  \textbf{9.00 / 26.47} &  9.15 / 26.52 &  9.36 / 27.00 \\
        & $1:9$ & 12.11 / 30.89 & 10.11 / 27.71 &  9.76 / 27.08 & 10.17 / 27.41 & 10.30 / 27.62 \\
        \midrule
        860hr & $1:4$ & 10.59 / 26.05 & 9.37 / 23.85 & 8.68 / 22.53 & 8.37 / 21.56 & \textbf{8.37} / \textbf{21.33} \\
        \bottomrule
    \end{tabular}
    \end{center}
\end{table*}

Table~\ref{tab:beam} shows how the WER varies with the beam size and the mixing ratio. 
For all mixing ratios, the model improves the most from a beam size of $k\!=\!1$ to $k\!=\!2$, showing the benefit of considering multiple hypotheses.
The improvement is greater when a higher mixing ratio of unpaired-to-paired speech is used.
In addition, we note that the LM is effectively unused when $k\!=\!1$ because the LM probability assigned to each hypothesis is normalized within the beam. If there is only one hypothesis, it will be assigned an approximate posterior probability of one.
The amount of improvement diminishes with larger beam sizes, and the performance even starts to degrade beyond $k\!=\!4$ when using a higher mixing ratio.
This may result from the inclusion of worse hypotheses which have a better score under the LM.
We use the best setting for the following experiments with a mixing ratio $m_{l} : m_{u} = 1 : 4$, a beam size $k\!=\!4$, and an LPM weight $\alpha\!=\!0.2$.
We present detailed results varying the LPM weight in the Supplementary Material.

\subsection{Proposal Model Update and Model Initialization}\label{sec:res_init_prop}

In addition to the two update strategies proposed in Section~\ref{sec:offpolicy}, termed \emph{On} and \emph{Off (better)}, we experiment with two additional strategies.
The first, \emph{Off (never)}, uses a fixed proposal model throughout training.
The second, \emph{Off (always)}, updates the proposal model with the online model every $T$ steps (i.e., set $\theta_r \leftarrow \theta_q$) regardless of performance.

The full results are shown in Table~\ref{tab:init_update}. 
Four key takeaways are as follows:
\begin{enumerate*}[label=(\arabic*)]
    \item For all combinations of $(\ryx, \qyx)$ initialization, off-policy (never) is the worst. This demonstrates the importance of updating the proposal model to generate better hypotheses during training.
    
    \item Off-policy (always) consistently outperforms on-policy. We observe that training is significantly stabilized by reducing the proposal model update frequency from every step to every 1,000 steps.
    The effect is particularly prominent when initializing $\ryx$ and $\qyx$ from an earlier checkpoint (9.62\% vs 20.02\% on dev-clean, and 27.51\% vs 45.62\% on dev-other).
    
    \item Off-policy (better) achieves the best WER in all settings and outperforms off-policy (always) by a larger margin when initializing from an earlier checkpoint.
    
    \item Unlike the other strategies, off-policy (better) demonstrates consistent improvement when using a less-trained initial online model.
\end{enumerate*}
In the following experiments, we initialize models with $\theta_r = \theta_A$ and $\theta_q = \theta_C$ unless otherwise specified.

\begin{table}[ht]
    \small
    \caption{We vary the proposal update strategies and the initial model weights for both $\ryx$ and $\qyx$.}
    \label{tab:init_update}
    \begin{center}
    \resizebox{\linewidth}{!}{%
    \begin{tabular}{llccc}
        \toprule
        \multirow{2}{*}{Init $\ryx$} & 
        \multirow{2}{*}{$\ryx$ update} & 
        \multicolumn{3}{c}{dev-clean / dev-other WERs} \\
        & & Init $\theta_q = \theta_A$ & Init $\theta_q = \theta_B$ & Init $\theta_q = \theta_C$ \\
        \midrule
        \multirow{4}{*}{$\theta_r = \theta_A$}
        & On             &  9.50 / 28.29 & N/A & N/A \\
        & Off (never)    & 11.19 / 31.74 & 11.14 / 31.69 & 11.24 / 31.53 \\
        & Off (always)   &  9.40 / 27.79 &  9.27 / 27.33 &  9.52 / 27.34 \\
        & Off (better)   &  9.20 / 27.42 &  9.14 / 26.80 &  \textbf{9.00} / \textbf{26.47} \\
        \midrule
        \multirow{4}{*}{$\theta_r = \theta_B$}
        & On             & N/A & 10.17 / 28.35 & N/A\\
        & Off (never)    & 13.61 / 35.39 & 13.95 / 35.43 & 13.56 / 35.62 \\
        & Off (always)   &  9.50 / 27.81 &  9.56 / 27.58 &  9.79 / 27.44 \\
        & Off (better)   &  9.30 / 27.34 &  9.26 / 27.01 &  \textbf{9.15} / \textbf{26.63} \\
        \midrule
        \multirow{4}{*}{$\theta_r = \theta_C$}
        & On             & N/A & N/A & 20.20 / 45.62 \\
        & Off (never)    & 20.59 / 44.09 & 22.95 / 46.43 & 23.42 / 46.89 \\
        & Off (always)   &  9.52 / 27.89 &  9.46 / 27.35 &  9.62 / 27.51 \\
        & Off (better)   &  9.44 / 27.34 &  \textbf{9.31} / 27.26 &  9.43 / \textbf{27.19} \\
        \bottomrule
    \end{tabular}
    }
    \end{center}
\end{table}

\subsection{Length Filtering}

Table~\ref{tab:filter} shows the impact of length filtering with different proposal model update strategies, where both the proposal model and the online model are initialized with $\theta_A$.
As discussed in Section~\ref{sec:filter}, on-policy suffers more than off-policy without length filtering.
If the proposal model is never updated, then length filtering does not affect the final WER. We hypothesize that length filtering keeps the proposal model stable during training.
In addition, we also compare three reference lengths: the oracle length, the predicted length from ASR + LM decoding, and that from ASR-only decoding. We observe that the WER does not differ much when using different reference lengths for filtering. Detailed results are shown in the Supplementary Material.

\begin{table}[ht]
    \small
    \caption{A comparison of WER with and without length filtering.}
    \label{tab:filter}
    \begin{center}
    \begin{tabular}{lcc}
        \toprule
        \multirow{2}{*}{$\ryx$ update} &  
        \multicolumn{2}{c}{dev-clean / dev-other WER} \\
        & No filtering & With filtering \\ 
        \midrule
        On-Policy             & 26.65 / 59.07 &  9.50 / 28.29 \\
        Off-Policy (never)    & 11.18 / 31.83 & 11.19 / 31.74 \\
        Off-Policy (always)   & 13.99 / 35.52 &  9.40 / 27.79 \\
        Off-Policy (better)   & 11.42 / 31.56 &  \textbf{9.20} / \textbf{27.42} \\
        \bottomrule
    \end{tabular}
    \end{center}
    \caption{Comparing $\py$ with different token perplexity.}
    \label{tab:lmppl}
    \begin{center}
    \begin{tabular}{lc}
        \toprule
        $\py$ PPL & dev-clean / dev-other WER \\
        \midrule
        34.24 &  \textbf{9.00} / \textbf{26.47} \\
        64.22 & 10.08 / 26.92 \\
        97.87 & 10.90 / 27.97 \\
        142.12 & 11.53 / 28.74 \\
        180.71 & 13.18 / 30.74 \\
        \bottomrule
    \end{tabular}
    \end{center}
\end{table}

\subsection{Choice of Language Models}

We study how the quality of the LM affects the results using the same ConvLM but trained for a varying number of steps.
We quantify LM quality with the token perplexity (PPL) on the development set.
Table~\ref{tab:lmppl} shows a clear positive correlation between the LM quality and the final WER. This is expected given that the better LM results in a more accurate posterior approximation, $\bpyx$.

\subsection{Comparison with Knowledge Distillation}
We next study different weak KD strategies and compare them with LPM. When multiple hypotheses ($k\!>\!1$) are used for KD, the student model matches against a uniform target distribution as done in~\citet{kahn2019self}. In addition, while KD typically considers a fixed teacher, we include another variant that adopts the same off-line teacher update strategy as LPM to disentangle the effect.
Results are shown in Table~\ref{tab:kd} with three key takeaways.
\begin{enumerate*}[label=(\arabic*)]
    \item Using an improving teacher leads to better performance.
    \item Distilling from a posterior that combines the ASR and LM models achieves better results; however, this hinders the use of an improving model as discussed in Section~\ref{sec:related}.
    \item Incorporating multiple hypotheses can slightly improve the performance even with a uniform target. Nonetheless, the gain is noticeably smaller than matching with our proposed local prior, which can be seen by comparing the ``(ASR, Imp, 4)'' rows, where the only difference is the target distribution to match.
\end{enumerate*}

\begin{table}[ht]
    \small
    \caption{Comparison with various knowledge distillation configurations. Each configuration is parameterized by (ASR/ASR+LM, Fix/Imp, $k$), where \emph{ASR/ASR+LM} indicates whether hypotheses are generated from ASR-only decoding or ASR+LM shallow fusion decoding, \emph{Fix/Imp} indicates whether the teacher model is fixed or improving (using our proposed off-policy update strategy), and $k$ indicates the number of hypotheses used for each utterance.}
    \label{tab:kd}
    \begin{center}
    \begin{tabular}{llc}
      \toprule
      Method & Param & dev-clean / dev-other WER \\
      \midrule
      \multirow{9}{*}{KD}
      & (ASR+LM,Fix,1) &  9.60 / 29.00 \\
      \cmidrule{2-3}
      %\midrule
      & (ASR,Fix,1)   &  12.24 / 33.25 \\
      & (ASR,Fix,2)   &  11.94 / 32.19 \\
      & (ASR,Fix,4)   &  11.60 / 32.26 \\
      & (ASR,Fix,8)   &  11.77 / 32.07 \\
      \cmidrule{2-3}
      %\midrule
      & (ASR,Imp,1)   &  11.79 / 30.09 \\
      & (ASR,Imp,2)   &  11.79 / 29.89 \\
      & (ASR,Imp,4)   &  12.09 / 30.21 \\
      & (ASR,Imp,8)   &  12.19 / 29.88 \\
      \midrule 
      LPM & (ASR,Imp,4) &   9.00 / 26.47\\
      \bottomrule
    \end{tabular}
    \end{center}
\end{table}

\subsection{Final Results and Comparison to Prior Work}
The best performing model is trained for 3.2M steps, with a learning rate annealed by a factor of two every 1.28M steps when using 360 hours of unpaired speech, and every 0.64M steps when using 860 hours of unpaired speech.
Reference lengths for filtering are obtained from ASR+LM beam search decoding.
We compare LPM to fully supervised models and a number of semi-supervised ASR techniques in Table~\ref{tab:main}.
Among the listed studies, pseudo labeling (PL) is the most comparable alternative as we follow the same experimental setup, and, more importantly, it achieved the previous state-of-the-art results on LibriSpeech when using train-clean-100 as paired data and train-other-360 as unpaired speech. 
We give a more detailed table comparing to prior work in the Supplemental Material.

The upper half of Table~\ref{tab:main} shows greedy decoding results without an LM. 
For the fully supervised model, when removing train-clean-360 the WER increases by 6.86\% on test-clean and 13.36\% on test-other.
Using train-clean-360 speech without transcripts, LPM reduces the absolute WER by 5.64\% and 12.21\% on the two test sets, which recovers 82\% and 91\%, respectively, of the WER drop from removing the labels.
Adding noisier train-other-500 to the unpaired set (total 860hr $\mathcal{D}_u^s$) further reduces the WER, and LPM achieves a better WER on the noisy sets (dev-other and test-other) compared to the supervised model trained on 460 hours of clean paired data.
In addition, LPM outperforms PL in all settings.
This trend is consistent even when decoding with a strong ConvLM.

The last row in the upper and lower halves of Table~\ref{tab:main} show the results of using LPM on the 60k hours of unlabelled speech from Libri-Light. We see that the WER improves by another 15\% and 9\% relative over using the 860hr dataset on the clean and other test sets respectively. When training on the 60k hours we use a batch size of 128, a beam size of $k\!=8\!$ and no learning rate decay. Reference lengths are from ASR-greedy decoding and we filter empty transcriptions, yielding 55.8k hours of training data. We also use a larger TDS model with the same architecture as~\citet{hannun2019sequence} (11 TDS blocks instead of 9). To disentangle the effect of the larger model from more unlabelled data with LPM, we also trained the larger model on the 860hr dataset. In this case, we did not see a gain in WER, suggesting that the improvement is due to LPM with more unlabelled data.

\begin{table*}[ht]
    \small
    \caption{Results of baselines and the proposed methods. Word error rate recovered (WERR) measures the improvment relative to the gap between supervised model trained on 100hr and 460hr of data, computed as $\text{WERR}(x)=(\text{WER}_{100hr} - x) / (\text{WER}_{100hr} -\text{WER}_{460hr})$. "-" means the number is not available, or the supervised results are not provided and hence WERR cannot be computed.}
    \label{tab:main}
    \begin{center}
    \resizebox{\linewidth}{!}{%
    \begin{tabular}{llllcccccc}
        \toprule
        & \multirow{2}{*}{$\mathcal{D}_{l}$} & \multirow{2}{*}{$\mathcal{D}_u$} & \multirow{2}{*}{LM}
        & \multicolumn{2}{c}{dev WER (\%)} & \multicolumn{2}{c}{test WER (\%)}
        & \multicolumn{2}{c}{test WERR (\%)}\\
        & & & & clean & other & clean & other \\
        \midrule
BT~\citep{hayashi2018back} & 100hr & 360hr (T) & None & 23.5 & - & 23.6 & - & 11.9 & - \\
Crit-LM~\citep{liu2019adversarial} & 100hr & 360hr (T) & None & 19.1 & - & 19.2 & - & - & - \\
Cycle-TTE~\citep{hori2019cycle} & 100hr & 360hr (S) & None & 21.5 & - & 21.5 & - & 27.6 & - \\
Cycle-TTS~\citep{baskar2019self} & 100hr & 360hr (S) & None & - & - & 17.9 & - & - & - \\
Cycle-TTS~\citep{baskar2019self} & 100hr & 360hr (S+T) & None & - & - & 17.5 & - & - & - \\
PL (ASR)~\citep{kahn2019self} & 100hr & 360hr (S)  & None & 12.27 & 33.42 & 12.57 & 35.36 & 33.24 & 34.36 \\
PL (ASR+LM)~\citep{kahn2019self} & 100hr & 360hr (S) + All (T) & None &  9.30 & 28.79 & 9.84 & 30.15 & 73.03 & 73.35 \\
PL (ASR+LM)~\citep{kahn2019self} & 100hr & 860hr (S) + All (T) & None &  9.03 & 26.03 & 9.44 & 27.25 & 78.86 & 95.06\\ 
\cmidrule{2-10}
Supervised & 100hr & N/A & None & 14.00 & 37.02 & 14.85 & 39.95 & 0.00 & 0.00 \\
Supervised & 460hr & N/A & None &  7.20 & 25.32 &  7.99 & 26.59 & 100.00 & 100.00 \\
Local Prior Matching & 100hr & 360hr (S) + All (T) & None &  8.85 & 26.33 & 9.21 & 27.74 & 82.22 & 91.39 \\
Local Prior Matching & 100hr & 860hr (S) + All (T) & None &  \hl{8.08} & \hl{21.52} & \hl{8.37} & \hl{22.89} & \hl{94.45} & \hl{132.19} \\
Local Prior Matching (large model)  & 100hr & 60,000hr (S) + All (T) & None & \bf{6.87} & \bf{19.92} & \bf{7.19} & \bf{20.84} & \bf{111.66} & \bf{143.04} \\
\midrule
PL (ASR)~\citep{kahn2019self} & 100hr & 360hr (S) & ConvLM & 6.19 & 23.53 & 6.81 & 24.99 & 32.64 & 41.66 \\
PL (ASR+LM)~\citep{kahn2019self}  & 100hr & 360hr (S) + All (T) & ConvLM & 5.73 & 22.54 & 6.35 & 24.13 & 44.65 & 48.24 \\
PL (ASR+LM)~\citep{kahn2019self}  & 100hr & 860hr (S) + All (T) & ConvLM & 6.31 & 21.87 & 6.84 & 23.29 & 31.85 & 54.66 \\
\cmidrule{2-10}
Supervised & 100hr & N/A & ConvLM & 7.78 & 28.15 & 8.06 & 30.44 & 0.00 & 0.00 \\
Supervised & 460hr & N/A & ConvLM & 3.98 & 17.00 & 4.23 & 17.36 & 100.00 & 100.00 \\
Local Prior Matching & 100hr & 360hr (S) + All (T) & ConvLM &  5.69 & 20.22 & 5.99 & 20.93 & 54.05 & 72.71 \\
Local Prior Matching & 100hr & 860hr (S) + All (T) & ConvLM & \hl{5.39} & \hl{14.89} & \hl{5.78} & \hl{16.27} &
\hl{59.53} & \hl{108.33} \\ 
Local Prior Matching (large model)  & 100hr & 60,000hr (S) + All (T) & ConvLM & \bf{4.87} & \bf{13.84} & \bf{4.88} & \bf{15.28} & \bf{83.03} & \bf{115.90} \\
        \bottomrule
    \end{tabular}
    }
    \end{center}
\end{table*}
\section{Analysis}

\subsection{Hypothesis Quality of Unlabeled Training Speech}
As discussed in Section~\ref{sec:res_init_prop}, updating the proposal model is crucial to improve the hypotheses used during training.
To quantify the improvement, Table~\ref{tab:labelqual} shows WERs on the unlabeled data of an LPM model at the beginning and at the end of training. Note that this is a proxy of quality for LPM, since multiple hypotheses are used when setting $k \ge 2$.
We compare this to the WER of pseudo-labels (PL) generated with an LM.
Although generating hypotheses without an LM is initially worse, as training progresses, the proposal model of LPM produces better predictions on both train-clean-360 and train-other-500 than the fixed ones used in PL.
Furthermore, the WER on train-other-500 is much higher for PL (21.51\%) than for LPM at the end of training (13.00\%), which explains why LPM achieves much better WER than PL when using the full 860hr of unpaired data.

\begin{table}[h]
    \small
    \caption{Comparison of the label quality on the unlabeled speech between PL and LPM when trained on 100hr $\mathcal{D}_{l}$ and 860hr $\mathcal{D}_u^s$.}
    \label{tab:labelqual}
    \begin{center}
    \begin{tabular}{llccc}
        \toprule
        \multicolumn{2}{l}{\multirow{2}{*}{Labelling Method}}
        & \multirow{2}{*}{Step} & \multicolumn{2}{c}{train WER (\%)}\\
        & & & clean-360 & other-500 \\
        \midrule
        PL    & ASR+LM stable     & All   &  8.25 & 21.51 \\
        LPM   & Proposal greedy   & 0     & 14.81 & 29.03 \\
        LPM   & Proposal greedy   & 3.2M  &  \textbf{7.37} & \textbf{13.00} \\
        \bottomrule
    \end{tabular}
    \end{center}
\end{table}

\subsection{Linguistic Plausibility}
We expect models trained with LPM to generate more semantically and grammatically correct text since the ASR model receives direct supervision from the LM.
Table~\ref{tab:decsample} shows the proposed hypotheses for two utterances using a supervised baseline model and a model trained with LPM.
The baseline model proposes erroneous hypotheses which are easy to discard even without the audio.
On the other hand, LPM generates hypotheses that are both grammatically and semantically plausible, with acceptable substitution errors in some cases (e.g., might/would).

We also notice in Table~\ref{tab:decsample} that the LM probabilities correlate well with linguistic plausibility for texts of similar lengths.
Motivated by this observation, we propose to quantify linguistic knowledge of an ASR model by measuring the LM perplexity of the hypotheses on the development set obtained using ASR-only greedy decoding.
Results are shown in Table~\ref{tab:decppl}. The ground truth text has the lowest perplexity on both sets as expected.
While all models are worse on dev-other than on dev-clean, LPM exhibits the smallest perplexity difference between the two sets, demonstrating that it successfully distills knowledge from the LM.

\begin{table}[h]
    \small
    \caption{LM token perplexity of ground truth texts and hypotheses obtained with greedy decoding.}
    \label{tab:decppl}
    \begin{center}
    \begin{tabular}{lllcccc}
        \toprule
        & \multirow{2}{*}{$\mathcal{D}_{l}$} & \multirow{2}{*}{$\mathcal{D}_u^s$}
        & \multicolumn{2}{c}{LM perplexity} \\
        & & & dev-clean & dev-other \\
        \midrule
        Ground Truth                    & N/A   & N/A   & 39.94 & 43.26 \\
        Supervised                      & 100hr & N/A   & 96.13 & 313.38 \\
        Supervised                      & 460hr & N/A   & 58.76 & 164.77 \\
        % \midrule
        PL (ASR)            & 100hr & 360hr & 87.36 & 273.14 \\
        PL (ASR+LM)         & 100hr & 360hr & 64.07 & 170.72 \\
        LPM                 & 100hr & 360hr & \textbf{61.73} & \textbf{159.72} \\
        LPM                 & 100hr & 860hr & \textbf{59.84} & \textbf{125.42} \\
        \bottomrule
    \end{tabular}    
    \end{center}
\end{table}
\section{Conclusion}
We introduce local prior matching, a semi-supervised learning objective for speech recognition, and demonstrate note-able reductions in WER with the addition of unpaired audio and text. We also perform an extensive empirical study to demonstrate the importance of various configurations of LPM.
While LPM is motivated by how humans learn to recognize speech, the proposed method can be applied to other sequence transduction tasks including machine translation~\citep{sennrich2015improving} and text summarization~\citep{nallapati2016abstractive}, provided a good prior for the domain. As LPM distills knowledge from a prior, it will be most effective when $\pxy$ is easy to model and $\py$ is more complex and hence difficult to learn with a limited amount of data.

We consider two promising directions for future work with LPM. First, we hypothesize that LPM can further benefit by incorporating more context in the prior, including previous sentences and signal from other modalities when available. Second, endowing the model with ability to dynamically select which examples to use for semi-supervision may further improve the effectiveness of LPM.

\section*{Acknowledgements}
The authors thank Jacob Kahn, Qiantong Xu, Tatiana Likhomanenko, Anuroop Sriram, Vineel Pratap, Vitaliy Liptchinsky, Ronan Collobert for their help and feedback.

\bibliography{main}
\bibliographystyle{icml2020}

\clearpage
\onecolumn
\icmltitle{Semi-Supervised Speech Recognition via Local Prior Matching -- Supplementary Materials}
\renewcommand\thetable{\thesection.\arabic{table}}
\appendix
\setcounter{table}{0}

\section{Additional Results}
Table~\ref{tab:alpha} shows the results of varying the LPM weight $\alpha$, as mentioned in Section 5.1. For this set of experiments, a mixing ratio $m_l:m_u = 1:4$ and a beam size $k=4$ is used.

\begin{table}[H]
    \small
    \caption{Results of varying LPM weight $\alpha$.}
    \label{tab:alpha}
    \begin{center}
    \begin{tabular}{lrr}
        \toprule
        \multirow{2}{*}{$\alpha$} & \multicolumn{2}{c}{dev WER (\%)} \\
        & clean & other \\
        \midrule
        2e-2 & 10.86 & 31.59 \\
        5e-2 & 10.08 & 28.92 \\
        1e-1 &  9.24 & 27.62 \\
        2e-1 &  \hl{9.00} & \hl{26.47} \\
        5e-1 &  9.41 & 26.56 \\
        \bottomrule
    \end{tabular}
    \end{center}
\end{table}

Table~\ref{tab:reflen} shows the LPM results when using different reference length estimates. As discussed in Section 5.3, the WER does not differ much when using different estimates, because we use the reference length to compute a \emph{range} for filtering for each utterance.
\begin{table}[H]
    \small
    \caption{Length filtering using different reference lengths.}
    \label{tab:reflen}
    \begin{center}
    \begin{tabular}{lc}
        \toprule
        Reference Length $L$        & dev-\{clean / other\} \\ 
        \midrule
        Oracle                      & 8.85 / 26.39 \\
        ASR + LM Dec                & \multirow{1}{*}{8.99 / 26.36} \\
        ASR-only Dec                & \multirow{1}{*}{9.00 / 26.47} \\
        \bottomrule
    \end{tabular}
    \end{center}
\end{table}

\clearpage
Table~\ref{tab:kd_ext} shows additional results of the comparison with knowledge distillation (KD) with varying initial teacher quality. In addition to the three key takeaways discussed in Section 5.5, the table here demonstrates that using an improving teacher can also reduce the sensitivity to its initial quality.

\begin{table}[H]
    \small
    \caption{Comparison with various knowledge distillation configurations. Each configuration is parameterized by (ASR/ASR+LM, Fix/Imp, $k$), where \emph{ASR/ASR+LM} indicates whether hypotheses are generated from ASR-only decoding or ASR+LM shallow fusion decoding, \emph{Fix/Imp} indicates whether the teacher model is fixed or improving (using our proposed off-policy update strategy), and $k$ indicates the number of hypotheses used for each utterance.}
    \label{tab:kd_ext}
    \begin{center}
    \begin{tabular}{llccc}
      \toprule
      \multirow{2}{*}{Method} & \multirow{2}{*}{Param} & 
      \multicolumn{3}{c}{dev-clean / dev-other WER (\%)} \\
      & & Init $\ryx=$ A & Init $\ryx=$ B & Init $\ryx=$ C \\ 
      \midrule
      \multirow{9}{*}{KD}
      & (ASR+LM,Fix,1) &  9.60 / 29.00 & 10.98 / 32.09 & 20.43 / 44.37 \\
      \cmidrule{2-5}
      & (ASR,Fix,1)   &  12.24 / 33.25 & 15.30 / 38.25 & 28.92 / 53.07 \\
      & (ASR,Fix,2)   &  11.94 / 32.19 & 14.63 / 36.85 & 26.81 / 49.32 \\
      & (ASR,Fix,4)   &  11.60 / 32.26 & 14.79 / 36.35 & 26.44 / 49.23 \\
      & (ASR,Fix,8)   &  11.77 / 32.07 & 14.48 / 36.50 & 26.81 / 49.59 \\
      \cmidrule{2-5}
      & (ASR,Imp,1)   &  11.79 / 30.09 & 14.29 / 32.10 & 18.81 / 35.84 \\
      & (ASR,Imp,2)   &  11.79 / 29.89 & 13.15 / 30.97 & 17.94 / 34.91 \\
      & (ASR,Imp,4)   &  12.09 / 30.21 & 13.46 / 31.22 & 16.64 / 34.56 \\
      & (ASR,Imp,8)   &  12.19 / 29.88 & 13.54 / 31.44 & 15.51 / 33.04 \\
      \midrule 
      LPM & (ASR,Imp,4) &   9.00 / 26.47 &  9.15 / 26.63 &  9.43 / 27.19 \\
      \bottomrule
    \end{tabular}
    \end{center}
\end{table}

Table~\ref{tab:main_cer} presents the character error rates (CERs) of the supervised models and the proposed methods.
\begin{table}[H]
    \small
    \caption{Character error rate (CER) results of the supervised models and the proposed methods.}
    \label{tab:main_cer}
    \begin{center}
    \begin{tabular}{llllcccc}
        \toprule
        & \multirow{2}{*}{$\mathcal{D}_{l}$} & \multirow{2}{*}{$\mathcal{D}_{u}^s$} & \multirow{2}{*}{LM}
        & \multicolumn{2}{c}{dev CER (\%)} & \multicolumn{2}{c}{test CER (\%)} \\
        & & & & clean & other & clean & other \\
        \midrule
        Supervised & 100hr & N/A & None & 6.20 & 20.27 & 6.80 & 22.14 \\
        Supervised & 460hr & N/A & None & 2.86 & 13.06 & 3.37 & 13.73 \\
        Local Prior Matching & 100hr & 360hr & None & 3.79 & 14.00 & 3.87 & 14.81 \\
        Local Prior Matching & 100hr & 860hr & None & 3.52 & 11.14 & 3.60 & 12.08 \\
        Local Prior Matching (large model) & 100hr & 60,000hr & None & 2.88 & 10.01 & 3.01 & 10.45   \\
        \midrule
        Supervised & 100hr & N/A & ConvLM & 3.83 & 17.03 & 3.86 & 18.52 \\
        Supervised & 460hr & N/A & ConvLM & 1.65 & 9.51 & 1.79 & 9.47 \\
        Local Prior Matching & 100hr & 360hr & ConvLM & 2.65 & 11.56 & 2.81 & 11.96 \\
        Local Prior Matching & 100hr & 860hr & ConvLM & 2.51 & 8.70 & 2.70 & 9.70\\ 
        Local Prior Matching (large model) & 100hr & 60,000hr & ConvLM & 2.32 & 7.61 & 2.19 & 8.53 \\
        \bottomrule
    \end{tabular}
    \end{center}
\end{table}

\clearpage
Table~\ref{tab:main_ext} shows the detailed results of semi-supervised ASR studies in the literature and the proposed methods. Word error rate recovered (WERR) for each baseline is computed using the supervised model WERs reported in its paper.
\begin{table}[H]
    \caption{A more comprehensive comparison with semi-supervised ASR studies using LibriSpeech, including the performances of the baseline/topline supervised model used in each study, since they differ significantly across different papers. 
    $\mathcal{D}_l$ and $\mathcal{D}_u$ denote the amount of paired and unpaired data used in each experiment, and (S)/(T)/(S+T) denote the type of the unpaired data, corresponding to speech/text/both, respectively. Experiments with the asterisk sign ($^*$) contain results that are not reported in the original paper, but are obtained from the authors of the paper.}
    \label{tab:main_ext}
    \begin{center}
    \resizebox{\linewidth}{!}{%
    \begin{tabular}{lllllrrrrrr}
        \toprule
        & & \multirow{2}{*}{$\mathcal{D}_{l}$} & \multirow{2}{*}{$\mathcal{D}_{u}$} & \multirow{2}{*}{LM}
        & \multicolumn{2}{c}{dev WER (\%)} & \multicolumn{2}{c}{test WER (\%)}
        & \multicolumn{2}{c}{test WERR (\%)}\\
        & & & & & clean & other & clean & other & clean & other \\
        \midrule
        \multirow{5}{*}{\citep{hayashi2018back}}
        & Supervised & 100hr & N/A   & None   & 24.9 & - & 25.2 & - & 0.0 & - \\
        & Supervised & 460hr & N/A   & None   & 11.4 & - & 11.8 & - & 100.0 & - \\
        & BT & 100hr & 360hr (T) & None       & 23.5 & - & 23.6 & - & 11.9 & - \\
        \cmidrule{2-11}
        & Supervised & 100hr & N/A   & RNN-LM & 23.0 & - & 22.9 & - & - & - \\
        & BT & 100hr & 360hr (T)     & RNN-LM & 21.6 & - & 22.0 & - & - & - \\
        \midrule
        \multirow{6}{*}{\citep{liu2019adversarial}}
        & Supervised & 100hr & N/A   & None     & 21.6 & - & 21.7 & - & - & - \\
        & Crit-LM & 100hr & 360hr (T) & None   & 19.1 & - & 19.2 & - & - & - \\
        & Crit-LM & 100hr & 860hr (T) & None   & 18.5 & - & 18.7 & - & - & - \\
        \cmidrule{2-11}
        & Supervised & 100hr & N/A   & RNN-LM   & 20.0 & - & 20.3 & - & - & - \\
        & Crit-LM & 100hr & 360hr (T) & RNN-LM & 17.1 & - & 17.3 & - & - & - \\
        & Crit-LM & 100hr & 860hr (T) & RNN-LM & 15.3 & - & 15.8 & - & - & - \\
        \midrule
        \multirow{5}{*}{\citep{hori2019cycle}}
        & Supervised & 100hr & N/A   & None & 24.9 & - & 25.2 & - & 0.0 & -  \\
        & Supervised & 460hr & N/A   & None & 11.4 & - & 11.8 & - & 100.0 & - \\
        & Cycle-TTE & 100hr & 360hr (S) & None & 21.5 & - & 21.5 & - & 27.6 & - \\
        \cmidrule{2-11}
        & Supervised    & 100hr & N/A       & RNN-LM & 22.6 & - & 22.9 & - & - & - \\
        & Cycle-TTE     & 100hr & 360hr (S) & RNN-LM & 19.6 & - & 19.5 & - & - & - \\
        \midrule
        \multirow{6}{*}{\citep{baskar2019self}}
        & Supervised    & 100hr & N/A         & None   & - & - & 21.0 & - & - & - \\
        & Cycle-TTS     & 100hr & 360hr (S)   & None   & - & - & 17.9 & - & - & - \\ 
        & Cycle-TTS     & 100hr & 360hr (S+T) & None   & - & - & 17.5 & - & - & - \\ 
        \cmidrule{2-11}
        & Cycle-TTS     & 100hr & 360hr (T)   & RNN-LM & - & - & 17.0 & - & - & - \\ 
        & Cycle-TTS     & 100hr & 360hr (S)   & RNN-LM & - & - & 16.8 & - & - & - \\ 
        & Cycle-TTS     & 100hr & 360hr (S+T) & RNN-LM & - & - & 16.6 & - & - & - \\ 
        \midrule
        \multirow{12}{*}{\citep{kahn2019self}}
        & Supervised        & 100hr & N/A         & None   & 14.00 & 37.02 & 14.85 & 39.95 & 0.00 & 0.00 \\
        & Supervised        & 460hr & N/A         & None   &  7.20 & 25.32 &  7.99 & 26.59 & 100.00 & 100.00 \\
        & PL (ASR)$^*$      & 100hr & 360hr (S)   & None   & 12.27 & 33.42 & 12.57 & 35.36 & 33.24 & 34.36 \\
        & PL (ASR+LM)       & 100hr & 360hr (S) + All (T)   & None   &  9.30 & 28.79 & 9.84 & 30.15 & 73.03 & 73.35 \\
        & PL (ASR+LM)$^*$   & 100hr & 860hr (S) + All (T)   & None   &  9.03 & 26.03 & 9.44 & 27.25 & 78.86 & 95.06 \\
        & PL (Ensemble)     & 100hr & 360hr (S) + All (T)   & None   &  8.60 & 27.78 & 9.21 & 29.29 & 82.22 & 79.79 \\
        \cmidrule{2-11}
        & Supervised        & 100hr & N/A         & ConvLM &  7.78 & 28.15 & 8.06 & 30.44 & 0.00 & 0.00 \\
        & Supervised        & 460hr & N/A         & ConvLM &  3.98 & 17.00 & 4.23 & 17.36 & 100.00 & 100.00 \\
        & PL (ASR)$^*$      & 100hr & 360hr (S)   & ConvLM &  6.19 & 23.53 & 6.81 & 24.99 & 32.64 & 41.66 \\
        & PL (ASR+LM)       & 100hr & 360hr (S) + All (T)   & ConvLM &  5.73 & 22.54 & 6.35 & 24.13 & 44.65 & 48.24 \\
        & PL (ASR+LM)$^*$   & 100hr & 860hr (S) + All (T)   & ConvLM &  6.31 & 21.87 & 6.84 & 23.29 & 31.85 & 54.66 \\
        & PL (Ensemble)     & 100hr & 360hr (S) + All (T)   & ConvLM &  5.37 & 22.13 & 5.93 & 24.07 & 55.47 & 48.70 \\
        \midrule
        \multirow{10}{*}{This work} 
        & Supervised    & 100hr & N/A   & None & 14.00 & 37.02 & 14.85 & 39.95 & 0.00 & 0.00 \\
        & Supervised    & 460hr & N/A   & None &  7.20 & 25.32 &  7.99 & 26.59 & 100.00 & 100.00 \\
        & LPM           & 100hr & 360hr (S) + All (T) & None &  8.85 & 26.33 &  9.21 & 27.74 & 82.22 & 91.39 \\
        & LPM           & 100hr & 860hr (S) + All (T) & None &  \hl{8.08} & \hl{21.52} & \hl{8.37} & \hl{22.89} & \hl{94.45} & \hl{132.19} \\
        & LPM           & 100hr & 60,000hr (S) + All (T) & None & \bf{6.87} & \bf{19.92} & \bf{7.19} & \bf{20.84} & \bf{111.66} & \bf{143.04} \\
        \cmidrule{2-11}
        & Supervised    & 100hr & N/A   & ConvLM &  7.78 & 28.15 & 8.06 & 30.44 & 0.00 & 0.00  \\
        & Supervised    & 460hr & N/A   & ConvLM &  3.98 & 17.00 & 4.23 & 17.36 & 100.00 & 100.00 \\
        & LPM           & 100hr & 360hr (S) + All (T) & ConvLM &  5.69 & 20.22 & 5.99 & 20.93 & 54.05 & 72.71 \\
        & LPM           & 100hr & 860hr (S) + All (T) & ConvLM &  \hl{5.39} & \hl{14.89} & \hl{5.78} & \hl{16.27} &
\hl{59.53} & \hl{108.33} \\
        & LPM           & 100hr & 60,000hr (S) + All (T) & ConvLM & \bf{4.87} & \bf{13.84} & \bf{4.88} & \bf{15.28} & \bf{83.03} & \bf{115.90} \\
        \bottomrule
    \end{tabular}
    }
    \end{center}
\end{table}

\end{document}